\documentclass{article}

\usepackage[nonatbib,preprint]{neurips_2021}

\usepackage[utf8]{inputenc} 
\usepackage[T1]{fontenc}    
\usepackage{hyperref}       
\usepackage{url}            
\usepackage{booktabs}       
\usepackage{amsfonts}       
\usepackage{nicefrac}       
\usepackage{microtype}      
\usepackage{graphicx}
\usepackage{color}
\usepackage{xcolor}
\usepackage{lipsum}
\usepackage[toc,page]{appendix}

\newcommand{\nl}{Relay Variational Inference}
\newcommand{\ns}{RVI}

\newcommand{\afina}{}

\title{\nl: \\ A Method for Accelerated Encoderless VI}

\author{%
  Amir Zadeh, Santiago Benoit, Louis-Philippe Morency \\
  Language Technologies Institute, Department of Computer Science\\
  Carnegie Mellon University\\
  Pittsburgh, PA 15213 \\
  \texttt{\{abagherz,sbennoit,morency\}@cs.cmu.edu} \\
}

\begin{document}

\maketitle

\begin{abstract}
Variational Inference (VI) offers a method for approximating intractable likelihoods. In neural VI, inference of approximate posteriors is commonly done using an encoder. Alternatively, encoderless VI offers a framework for learning generative models from data without encountering suboptimalities caused by amortization via an encoder (e.g. in presence of missing or uncertain data). However, in absence of an encoder, such methods often suffer in convergence due to the slow nature of gradient steps required to learn the approximate posterior parameters. In this paper, we introduce Relay VI (RVI), a framework that dramatically improves both the convergence and performance of encoderless VI. In our experiments over multiple datasets, we study the effectiveness of \ns \ in terms of convergence speed, loss, representation power and missing data imputation. We find \ns \ to be a unique tool, often superior in both performance and convergence speed to previously proposed encoderless as well as amortized VI models (e.g. VAE). \afina
\end{abstract}
\section{Introduction \afina}

Variational Inference (VI) offers a method for sampling from interactable posteriors in machine learning~\cite{jordan1999introduction}. It relies on using approximate posteriors: essentially well-known distributions that make the process of sampling from the latent space possible. Commonly in VI, inference is done using parametric models~\cite{dayan1995helmholtz}, such as a probabilistic encoder in VAE~\cite{kingma2013auto}. Using a parameteric model to perform inference allows for fast inference given new datapoints. However, encoders can also produce suboptimal approximations of the posteriors~\cite{cremer2018inference}, especially when inference is done over uncertain inputs, e.g. with missing data~\cite{zadeh2019variational}. Such suboptimalities in the encoder can in turn lead to suboptimal optimization of the variational objective (Evidence Lower Bound - ELBo). 

Inference of the approximate posterior parameters can be done optimally without relying on an encoder; using stochastic gradient-based optimization of the approximate posterior parameters directly~\cite{zadeh2019variational}. Encoderless models offer an intriguing perspective of VI, one that is robust to presence of noise or missing data, as well noise distribution disparity between train and test stages. However, their main shortcomings is their reliance on slow incremental optimization in inference of approximate posterior parameters. This slow nature is further intensified by the fact that approximate posterior parameters for the datapoints are optimized independently, due to the mean-field assumption as part of encoderless VI\footnote{as opposed to encoder-based inference where updates incrementally make the encoder better at posterior approximation for all the datapoints.}.

In this paper, we introduce \nl \ (\ns). \ns \ is a general (blackbox) neural VI framework that utilizes ``relays'' within the latent space to speed up the optimization of the variational objective. The relays essentially exploit the commonalities between datapoints according to their formulation (Section \ref{sec:model}). Subsequently, the gradient of the objective can flow between datapoints, essentially mitigating the slow nature of mean-field approximation. This in turn decreases the number of gradient steps during training and inference, thus increasing the overall speed. We summarize the characteristics of the proposed method as follows: 

\begin{itemize}
    \item \ns \ maintains the appealing properties of encoderless VI, including optimal inference as well as robustness to noisy or missing data, while speeding up the training and inference process. 
    
    \item The benefits of \ns \ come without requiring extra sampling steps, loss terms, or arduous increase in number of learnable parameters. 
    
    \item It is also simple to implement, and flexible in the manner in which the relay can be calculated and interpreted. 
\end{itemize}

\section{Related Works \afina}\label{sec:related}

In deep learning, inference of approximate posteriors in VI is commonly done with encoders (i.e. inference networks). While amortizing the posterior inference has benefits when it comes reducing the time required for this crucial step of VI, it can also suffer from suboptimalities~\cite{cremer2018inference}, especially in presence of uncertain input (e.g. missing data)~\cite{zadeh2019variational}. Previous works have attempted to mitigate the inference suboptimality of encoders using fine-tuning~\cite{hjelm2016iterative}, ladder-based models~\cite{sonderby2016ladder}, or Hessian-based finetuning after initial training~\cite{kim2018semi}. While designed for particular architectures (e.g. RNNs), or limited by the maximum number of learnable parameters (due to Hessian calculations), generalizable extensions of such models require further research. 

More relevant to this paper, are the works that rely on updating the approximate posteriors using a gradient-based optimization pipeline~\cite{hoffman2013stochastic,zadeh2019variational}. Such models directly optimize the parameters of the posterior distribution, which can mitigate the suboptimalities of an amortized framework. They can also remain robust in presence on noise or missing data. However, the updates of such gradient-based posterior approximation frameworks can be slow and unsteady~\cite{dhaka2020robust}. They remain susceptible to the choice of training parameters such as learning rate. A common cause of the slow and unsteady nature of their training is the full independence (mean-field assumption) imposed by the gradient-based posterior approximation methods~\cite{hoffman2015structured}. Making strong assumptions about the hierarchy in the latent space and enforcing extra loss terms on the VI is proposed to mitigate this issue~\cite{ranganath2016hierarchical}. However, such strong assumptions may introduce unnecessary variances on both the latent space and the learning framework. 

Through usage of relay, the method in this paper proposes a way to break the mean-field assumption, without requiring strong assumptions on the latent space (e.g. through new loss terms in the VI objective). Furthermore, the relays can take many flexible configurations to offer interpretable concepts such as hierarchies and clusters.

\section{Model}\label{sec:model}

Let $\mathcal{D}=\{x_i \sim p(x);\ x \in \mathbb{R}^{d}\}_{i=1}^N$ be a given datasets. $d$ denotes the dimensionality of individual datapoints. Given this dataset, the goal is to learn a parametric distribution $p_\theta(x)$, with learnable parameters $\theta$. The generative process of the dataset is considered as $p(z)p(x|z)$, where a sample of the latent space $z_i \sim p(z)$ is used to create the datapoint $x_i$ using the conditional distribution $p(x|z)$. However, this would require calculation of the true posterior $p(z|x)$ during train time, which is often hard to calculate for nonlinear decoders without MCMC. 

Instead of true posterior, an approximate posterior $q$ can be marginalized over during training, with the condition that $q(z|x_i, \phi)>0$ if $p(z|x_i,\theta)>0$. Using this approximate posterior, the likelihood of the model parameters given the data can be written as:

\begin{equation}\label{eq:likelihood}
    \mathcal{L}(\theta | x_i)=\textrm{KL}\Big(q_\phi(z|x_i) \ \Big|\Big| \ p_\theta(z|x_i)\Big) + \mathcal{V} (q_\phi,\theta|x_i)
\end{equation} 

$\mathcal{V}(\cdot)$, often referred to as ELBo, can be written as:
\begin{equation}\label{eq:elbo}
    \mathcal{V} (q_\phi,\theta|x_i)=\mathbb{E}_{q_\phi(z|x_i)} \Big[\textrm{log}\ p_\theta(x|z)\Big] - \textrm{KL} \Big(q_\phi(z)\ \Big|\Big|\ p_\theta(z)\Big)
\end{equation}

In presence of missing data, the first term of the RHS in Equation \ref{eq:elbo} marginalizes over the missing dimensions \cite{zadeh2019variational}.

In \nl \ framework, the parameters of the local approximate posterior distributions can be written as:

\begin{equation}\label{eq:phi}
    \phi_i=\phi_i^R + \phi_i^{\epsilon}
\end{equation}

In the above formulation, $\phi_i^R$ is called the relay. It is a factor that binds a set of datapoints together thus breaking the mean-field assumption of the stochastic optimization process. It's main role is to and allow for propagation of the gradient among the datapoints. The formulation of the relay can take different forms, and is discussed in the next section. $\phi_i^{\epsilon}$ is a stochastic variable that is unique to individual datapoints. 

\subsection{Formulation of $\phi_i^R$ } \label{sec:relay}

The relay $\phi_i^R$ is essentially a shared factor among the datapoints. The calculation of $\phi_i^R$ can take different forms\footnote{A comprehensive set of such forms with their charactristics are discussed in supplementary, including mixture models and clusters.}, but in this section we discuss one particular approach that offers fast convergence without sacrificing the learning performance. Let $R=\{r_k \in \mathbb{R}^{t}\}_{k=1}^K$ be a set of $K$ independent stochastic vectors within the latent space with $t$ dimensions. These $K$ vectors can in turn be used to approximate the position of a given data point within the latent space. This is formulated as follows:

\begin{equation}\label{eq:relay}
    \phi_i^R=\Sigma^{b_i \in K} \ a_{b_i} \ \cdot \ r_{b_i}
\end{equation}

Where $B_i=\{b_i \in K\}$ are indicators for the different vectors in $R$, and $a_{b_i}$ are coefficients for vectors $r_{b_i}$. In simple terms, $i$th datapoint is assigned to a subset of vectors $\{b_i \in K\}$ with coefficients $a_{b_i}$. The relay is a direct linear combination of the subset as shown in Equation \ref{eq:relay}, which in turn defines the position of the approximate posterior parameters in the latent space, according to the Equation \ref{eq:phi}. 

In reality, without complicating the notation\footnote{Full notation and calculation provided in supplementary.}, there can be multiple sets of vectors $R$, defining multiple groups of vectors which a datapoint is defined based on. For example, a first small set can be defined with only a few vectors to capture the holistic variations among datapoints, while a subsequent group with larger number of vectors can capture the differences across the datapoints. The $\phi_i^R$ is calculated for each of the different groups according to Equation \ref{eq:relay}. The calculated relays based on each group can subsequently be summed to calculate the position of a final relay which will be used in Equation \ref{eq:phi}. In most of our experiments, we use the combination of three groups with $25$, $50$ and $100$ vectors. Each datapoint is assigned to only a total of $|B_i|$ vectors in each group ($|B_i|$ can change from one group to another). In each group, the choice of which vectors will be assigned to the datapoint is done by first calculating the coefficients for all vectors, and subsequently choosing them based on the top $|B_i|$ coefficients according to their absolute value (e.g. top $10$ absolute values in a group of $50$ vectors). 

\section{Learning}
Learning can be done by maximizing the lower bound in Equation \ref{eq:elbo}. Using Equations \ref{eq:phi} and \ref{eq:relay}, Equation \ref{eq:elbo} is differentiable w.r.t $R$ (or multiple $R$s if there exist multiple groups), $a_{b_i}$ and $\phi_i^{\epsilon}$. All these learnable variables together define the parameters of the approximate posterior for each data point. They can be initialized randomly, and optimized as stochastic variables using gradient-based methods, alongside the parameters $\theta$ of the decoder. After training is done, for new datapoints, Equation \ref{eq:elbo} is used to optimize $a_{b_i}$ (assignment coefficients) and $\phi_i^{\epsilon}$. $R$ and $\theta$ are the parameters of the model, and remain fixed during testing. 

Since \ns \ introduces no new distribution or loss term to the lower bound in Equation \ref{eq:elbo}, sampling during generation remains the same as conventionally done in literature \cite{kingma2013auto}.

While learning is done using the ELBo objective, to have a universally understandable sense of model performance in our experiments, we report the elastic metric (L1+L2, lower better) for the reconstruction done based on drawing the mean of the approximate posterior distributions. 
\section{Methodology \afina}\label{sec:method}

In this section, we outline the methodology for the experiments of this paper. We first discuss the datasets and baselines, followed by a discussion of the network architecture used in our experiments.

\subsection{Datasets and Baselines}

\textbf{MNIST and Fashion-MNIST:} MNIST and Fashion-MNIST\footnote{\url{https://github.com/zalandoresearch/fashion-mnist}} are datasets of $28\times28$ grayscale images which contain $10$ classes of digits or fashion items. We use the standard folds of the two datasets. 

\textbf{CMU-MOSI Dataset:} CMU Multimodal Sentiment Intensity (CMU-MOSI) is an in-the-wild dataset of multimodal sentiment analysis~\cite{zadeh2018multimodal}. The dataset contains $2199$ opinionated sentence utterances, with three modalities of text (words), vision (gestures) and acoustic (sound). The labels are sentiment scores between $[-3,3]$. Feature extraction on this dataset is similar to~\cite{chen2017multimodal}. We use the standard released folds\footnote{https://github.com/A2Zadeh/CMU-MultimodalSDK}. 

\textbf{Toy Artificial Dataset:} We use a similar artificial dataset as used by ~\cite{zadeh2019variational}. The input space of this dataset is $300$ dimensions. It is generated based on linear combination of samples drawn from $10$ different distributions such as Normal, Beta and Gumbel.

The following main baselines are used for comparison in our experiments. Further baselines in Section \ref{sec:related} are studied in the supplementary. 

\textbf{VAE:} Variational AutoEncoder (VAE) \cite{kingma2013auto} is an amortized neural VI model. It relies on an encoder to approximate the posterior distribution conditioned on a given datapoint. During training, the parameters of the encoder are learned alongside the decoder in an end-to-end fashion. 

\textbf{VAD:} Variational AutoDecoder (VAD) \cite{zadeh2019variational} is a VI method specifically useful for learning from missing data. Unlike a VAE which relies on an encoder to approximate the posterior of the data, VAD does so using an iterative optimization process over the ELBo loss (with the missing dimensions marginalized).

\begin{figure*}[t]
\centering{
\includegraphics[width=\linewidth]{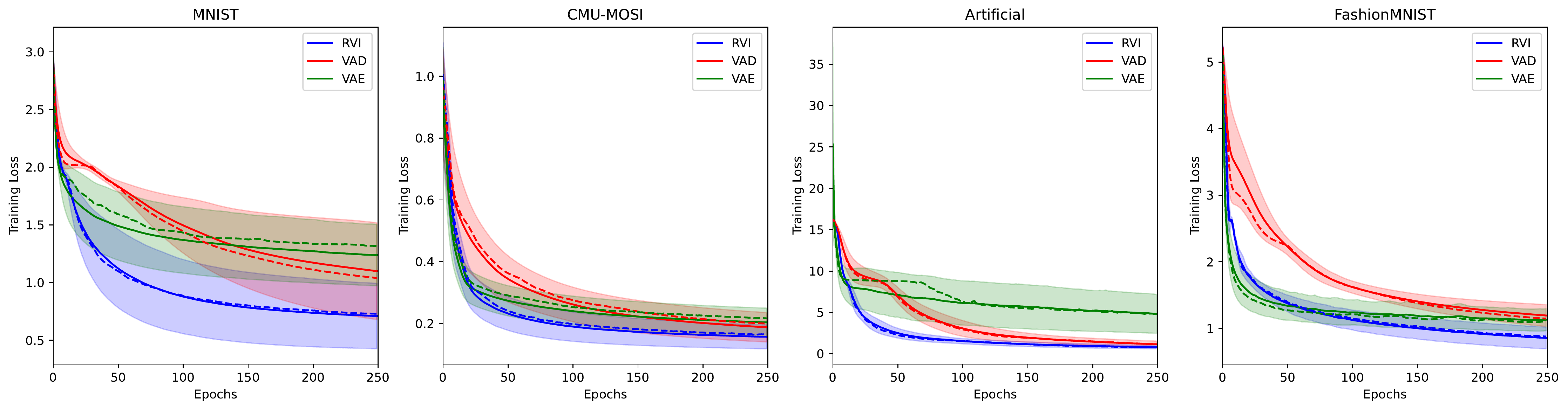}
\caption{\label{fig:massgs} Convergence analysis of the compared models in this paper, across a range of hyperparameters discussed in Section \ref{sec:exp_train_convergence}. Dotted line displays the median performance. Overall, \ns \ (blue) maintains superior average loss than both VAE (green) and VAD (red). \afina }}
\end{figure*} 

\subsection{Neural Decoder Architecture}\label{sec:decoder}

In this paper, we perform our experiments using three different neural architectures. The main decoder architecture uses two hidden layers of $[64,64]$ neurons between the latent space ($64$ neurons) and the output space (same dimension as the input space). Hidden layers are activated using ReLU, with the final layer having linear activation. For VAE, the encoder is the reverse architecture of the decoder. Two auxiliary architectures are also studied in Section \ref{sec:experiments} and more in depth in the supplementary\footnote{We also study the effect of latent space shape in the supplementary.}. These architectures are one-layer decoder with $[64]$ hidden units, and three-layer decoder $[64,64,64]$ hidden units. Similarly, they use ReLU as hidden layer activation and linear as final layer. Overall, in our experiments we observe closely similar baseline performance trends for the auxiliary architectures and the main architecture.

\section{Experiments}\label{sec:experiments}
In this section, we outline a comprehensive set of experiments to understand the performance of \ns \ in detail. We recommend viewing the figures in color, and zoomed in if necessary. We recommend viewing the figures in color, and zoomed in if necessary. We put the primary focus on the first term of the variational inference for our experiments. The first term is what experiences instability due to missing values, with the second term being identical across different variational baselines in this paper. 

\begin{figure*}[t]
\centering{
\includegraphics[width=\linewidth]{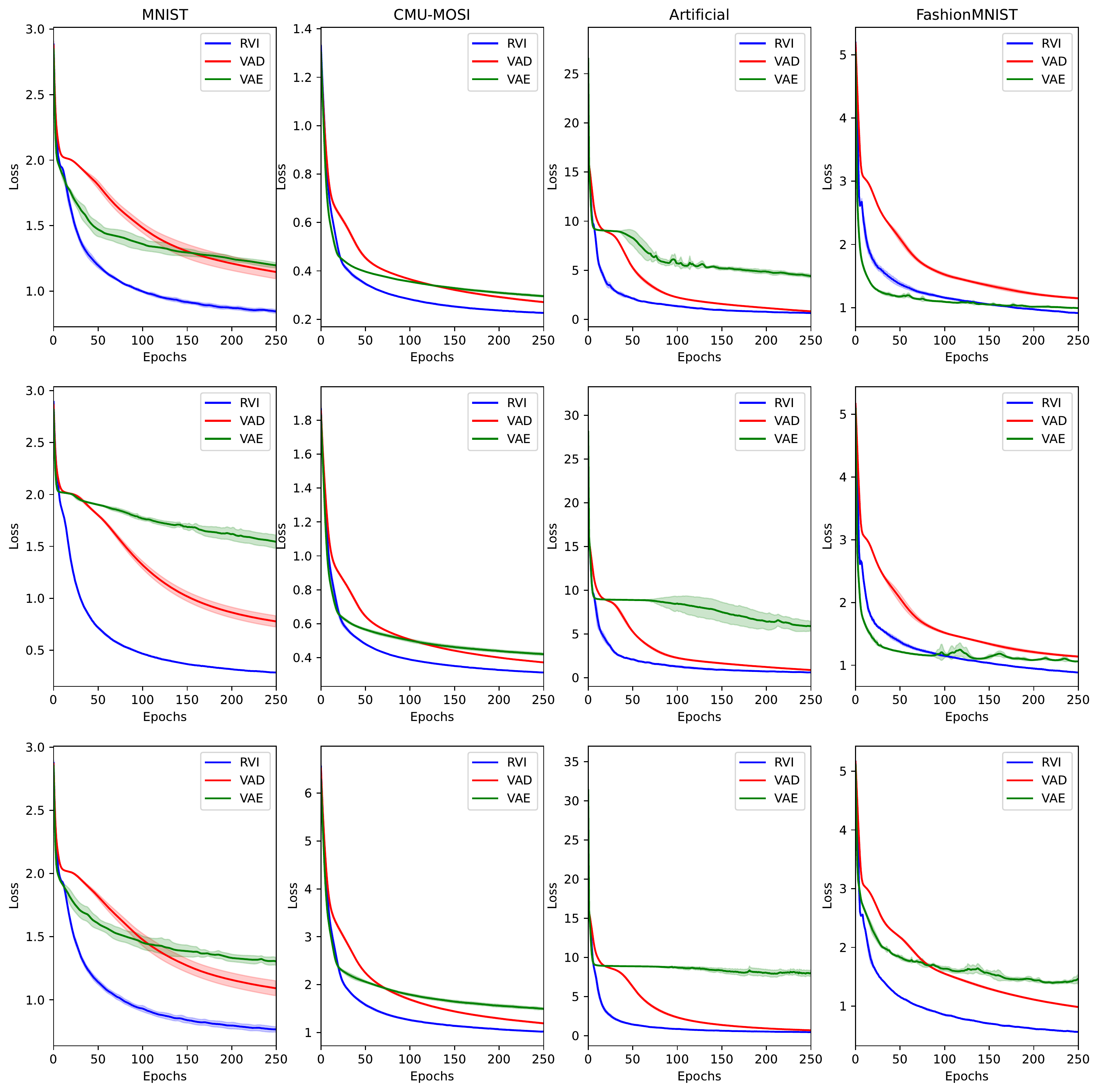}
\caption{\label{fig:convergence_train} Convergence comparison between the proposed \ns \ (blue) framework, VAD (red), and VAE (green) applied on the main network discuss in Section~\ref{sec:experiments}. Training is  performed using the commonly used learning rate of $0.001$ for Adam. First, second and third row are for $0.0$, $0.5$, and $0.9$ missing rates respectively. Generally, \ns \ converges faster, and to a better loss than baselines. }}
\end{figure*} 

\subsection{Training Convergence}\label{sec:exp_train_convergence}
We compare the convergence rate of the \ns, VAE, and VAD during training. This convergence analysis is performed over an extensive hyperparameter search space. The hyperparameters include the learning rates of $\{0.01,0.001,0.0001,0.00001\}$ (for both the network $\theta$ and approximate posterior parameters $\phi_i^R$ and $\phi_i^{\epsilon}$), $3$ different decoder architectures $[64]$, $[64,64]$, $[64,64,64]$ (encoder is the inverse of the decoder for VAE), and Missing Completely at Random (MCAR) missing rates of $\{0.1,0.2,0.3,0.4,0.5,0.6,0.7,0.8,0.9\}$. Each hyperparameter set is trained $10$ times. Figure \ref{fig:massgs} shows the mean, standard deviation and median of the elastic metric for all the hyperparameter runs. It highlights the average behavior of the models over the entire set of trained hyperparameters. The results of this experiment demonstrate that \ns \ is able to perform superior and converge fast than both VAE and VAD. 

Aside a holistic view of performance across different hyperparameters, we also study how the models perform using the commonly used learning rate of $0.001$ for Adam \cite{kingma2014adam} (across all parameters $\theta$, $\phi_i^R$ and $\phi_i^{\epsilon}$). The decoder is the main $2$ layer network discussed in Section \ref{sec:decoder}. Figure \ref{fig:convergence_train} compares the 3 models side by side. The rows demonstrate the MCAR missing rates of $0.0$ (no missing data), $0.5$, and $0.9$. The columns are the datasets used in the experiment. Overall, \ns \ shows superior performance than VAE and VAD.

\begin{figure*}[t]
\centering{
\includegraphics[width=\linewidth]{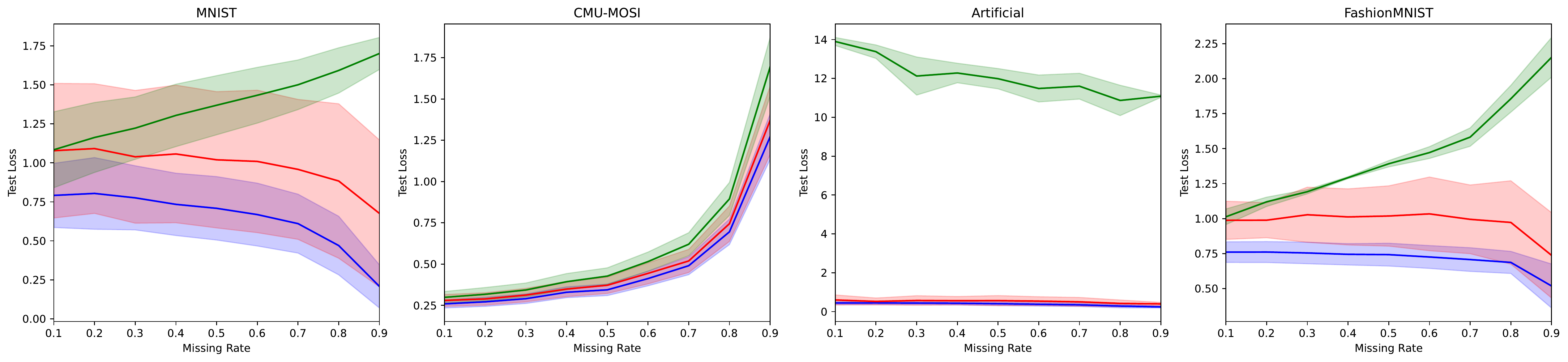}
\caption{\label{fig:testloss} Comparison of \ns \ (blue), VAE (green) and VAD (red) over different MCAR missing rates at test time. The loss is calculated over the available (non-missing) portions that are used for the inference. \ns \ outperforms both VAE and VAD in generalization to test set. }}
\end{figure*} 
\begin{figure*}[t]
\centering{
\includegraphics[width=\linewidth]{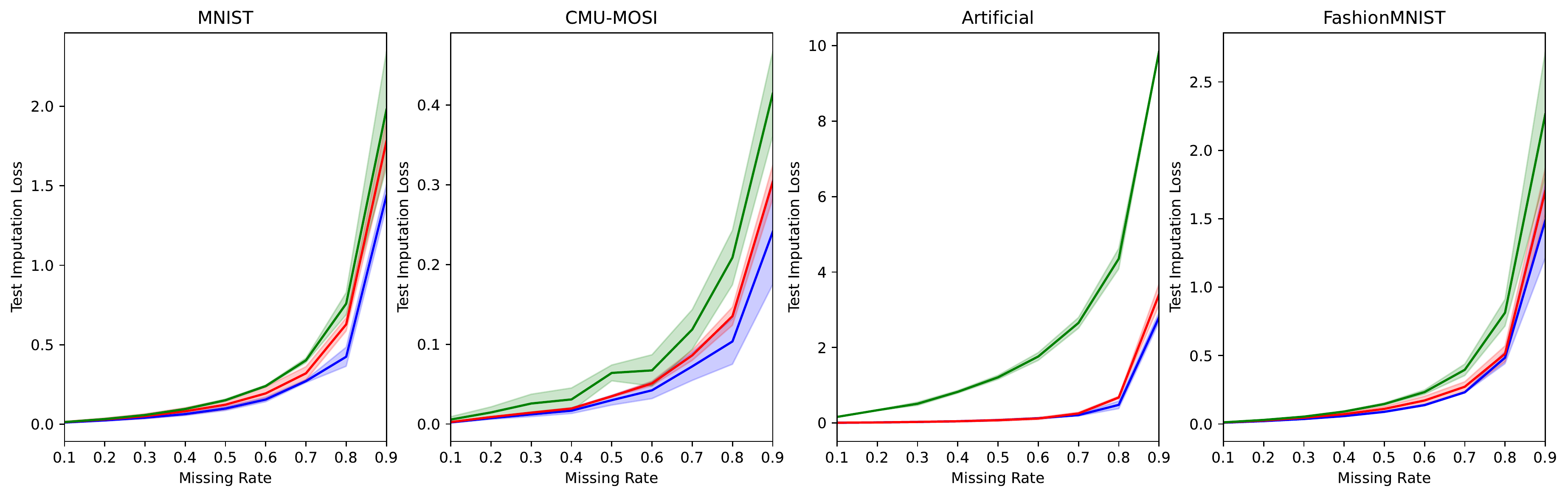}
\caption{\label{fig:imputation} Comparison of \ns \ (blue), VAE (green) and VAD (red) for missing data imputation at test time over different MCAR missing rates. After inference, the missing data is revealed to calculate the loss (thus missing components are never used during inference). This shows that while \ns \ is faster than VAD, it can also achieve similar or superior performance in imputation.}}
\end{figure*} 

\subsection{Test Performance}

We compare the models in this paper for the generalization of their performance over the test set. The testing is performed after the models are trained for a full $250$ epochs. The missing rate studied in this case are MCAR with rates of $\{0.1,0.2,0.3,0.4,0.5,0.6,0.7,0.8,0.9\}$. Figure \ref{fig:testloss} shows the test loss for the compared models over different missing rates. Standard deviations are calculated over $10$ runs of each hyperparameter set. We also extend this experiment to the task of missing data imputation. After inference is done using \ns, VAD or VAE, we reveal the missing components of each datapoint to calculate the imputation loss. Figure \ref{fig:imputation} shows the imputation performance of the compared models. 

Further studies of the latent space (including interpolations and visualizations), as well as results for other missing patterns are covered in supplementary.

\begin{figure*}[t]
\centering{
\includegraphics[width=.75\linewidth]{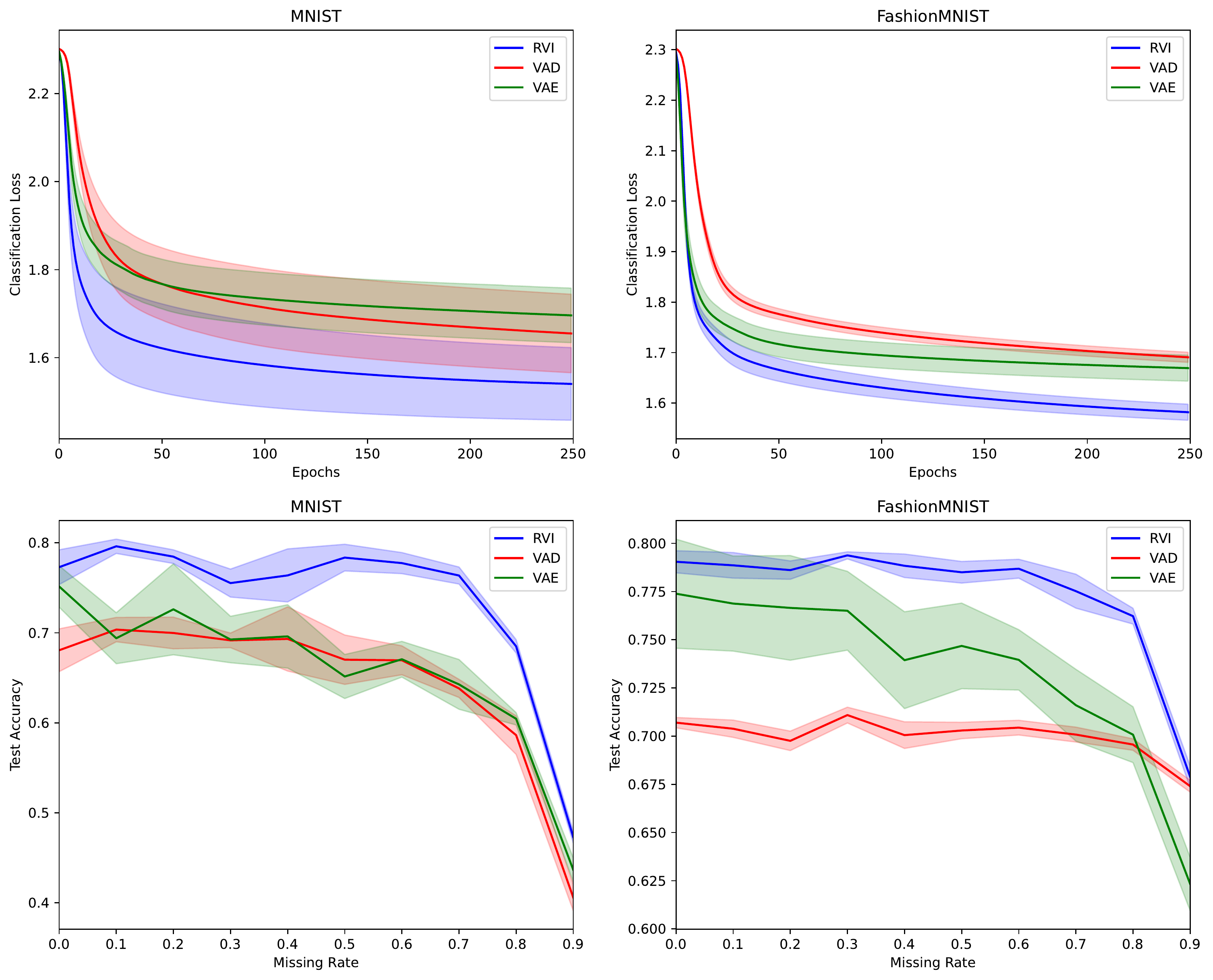}
\caption{\label{fig:sup_convergence} Comparison of convergence speed and accuracy of trained models over representations learned by \ns \ (blue), VAE (green) and VAD (red). The representations learned by \ns \ offer faster downstream supervised convergence and more accurate predictions. }}
\end{figure*} 
\subsection{Supervised Analysis}

While comparing the learned approximate posteriors in terms of how well their reconstructions resemble the datapoints is important, we also study the strength of the learned representations when used for supervised tasks. We first learn the representations from a dataset using the variational objective in Equation \ref{eq:elbo}, and subsequently use the representations ($\phi_i$) as input to a supervised model. For learning representations, we use the main network in Section \ref{sec:decoder}. We use the MNIST and the more challenging Fashion-MNIST variant, with their standard folds. We change the missing rate from $0.0$ to $0.9$ in increments of $0.1$. Representations are are trained for $250$ epochs, and subsequently used to train a supervised model with $64$ hidden neurons (ReLU activated) for $250$ epochs. Training is done $10$ times to calculate standard deviations. Learning rate for all networks and approximate posteriors is $0.001$. Figure \ref{fig:sup_convergence} shows the results of this experiment. The first row shows the classification loss (categorical cross-entropy) when training the classification network, and the second row shows the accuracy over test set. Overall, the representations learned by \ns \ offer faster convergence and more accurate predictions across different learning rates.

\subsection{Posterior Approximation Learning Rate}

\begin{figure*}[t]
\centering{
\includegraphics[width=0.65\linewidth]{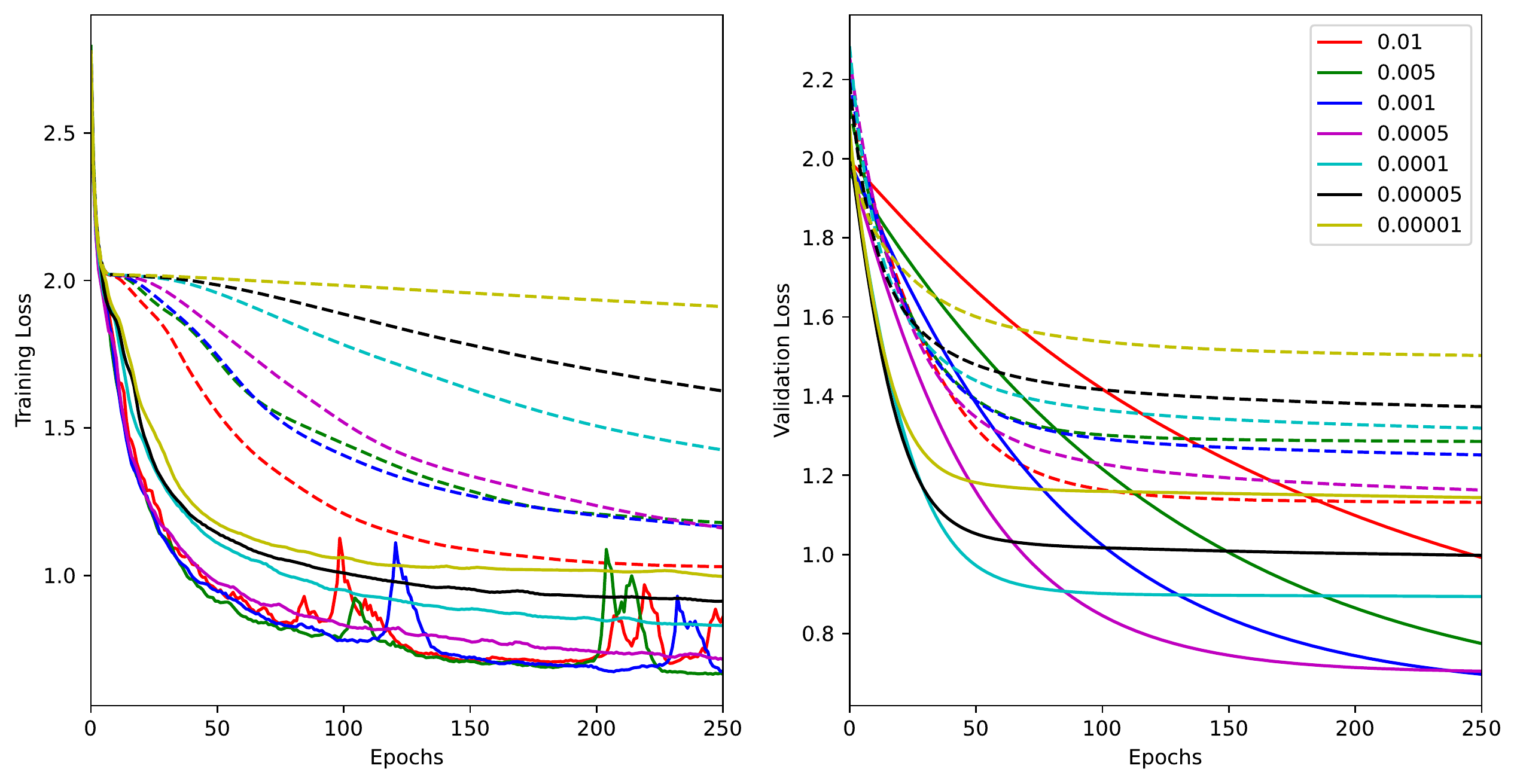}
\caption{\label{fig:difflr} The effect of the approximate posterior learning rate over training speed for \ns \ (solid) and VAD (dotted). MNIST dataset is used, and network learning rate is $0.001$. Learning rate search space includes $7$ different learning rates color coded in the figure. }}
\end{figure*}

While in the previous experiments, the learning rate of the network and the approximate posteriors are the same, in reality, the approximate posterior can have a different learning rate than the network. This change in learning rate can possibly affect the convergence, and needs to be studied for both \ns \ and VAD (VAE has no such hyperparameter therefore excluded from this experiment). We use the main network in Section \ref{sec:decoder}, with the learning rate of $0.001$ for $\theta$. The approximate posterior learning rates is chosen from $\{0.01, 0.005, 0.001, 0.0005, 0.0001, 0.00005, 0.00001\}$. The dataset is MNIST with no missing data. We study the convergence rate for both train and test sets in Figure \ref{fig:difflr}. We observe that during both training and testing, \ns \ achieves better loss than the VAD. The results also signal that for the best convergence, it is better to use high learning rate during training of \ns \ but use a lower one during validation. 

\subsection{Relay Structure}

\begin{figure*}[t]
\centering{
\includegraphics[width=.7\linewidth]{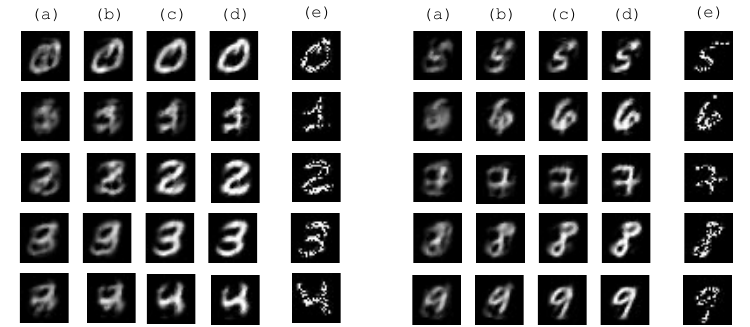}
\caption{\label{fig:progression} Reconstruction progression of the proposed \ns \ framework. $(\texttt{e})$ is the input to the \ns \ model, and $(\texttt{d})$ is the final reconstruction. $(\texttt{a})$ is the reconstruction using only the first group of $25$ relays, $(\texttt{b})$ is the contribution of the two groups of $\{25,50\}$ and $(\texttt{c})$ is the contribution of all groups $\{25,50,100\}$. $(\texttt{d})$ includes the contribution from $\phi_i^{\epsilon}$.}}
\end{figure*}

We study the effect of the relay structure, as well as the effect of the different groupings discussed in Section \ref{sec:relay}. The models are trained on MNIST dataset. The network used is the main network discussed in Section \ref{sec:decoder}. Learning rate of $0.001$ is used for both the network and the approximate posteriors. There are three groups of relays as defined in Section \ref{sec:relay} which include $\{25,50,100\}$ vectors each. Figure \ref{fig:progression} shows the progressive contribution of each of the groups to the final reconstruction of the input images (at MCAR rate of $0.5$). The results show that the relays in each group contribute to the final reconstruction of the datapoints. Furthermore, the group with $25$ vectors captures the general appearance of the digit with $50$ and $100$ capturing further variations needed for reconstructing the datapoints.


\section{Conclusion}

In this paper, we presented the \nl \ (\ns) framework. Using relays, \ns \ is able to bind datapoints together and mitigate the mean-field assumption required for encoderless VI. The relay is designed to improve the performance and convergence rate of encoderless VI. We perform experiments over multiple datasets, and compare the \ns \ to VAE and VAD. \ns \ reaches better loss, and in most cases converges faster than both VAE and VAD. It's performance generalizes to missing data imputation, performing similar or better than VAD. Furthermore, the representations learned by \ns \ are studied for supervised learning, and show to be superior to VAE and VAD. 

\clearpage
\bibliographystyle{plain}
\bibliography{citations}
\clearpage

\begin{appendices}
\section{Relay Formulations}\label{app:formulation}
The formulation of relay $\phi_i^R$ can take many different forms. However, in this paper we study two cases. The first is discussed in the main body of the paper. The second case is the case of assigning the datapoints to only a single vector in each relay group (essentially $|B_i|=1$, for each relay group). This essentially forms clusters of datapoints in the latent space, where each datapoint is assigned to one relay. For this, similar to the experiments in the main body of the paper, we utilize 3 groups of $\{25,50,100\}$, and compare it with the main formulation in the paper. 
\begin{figure*}[t]
\centering{
\includegraphics[width=\linewidth]{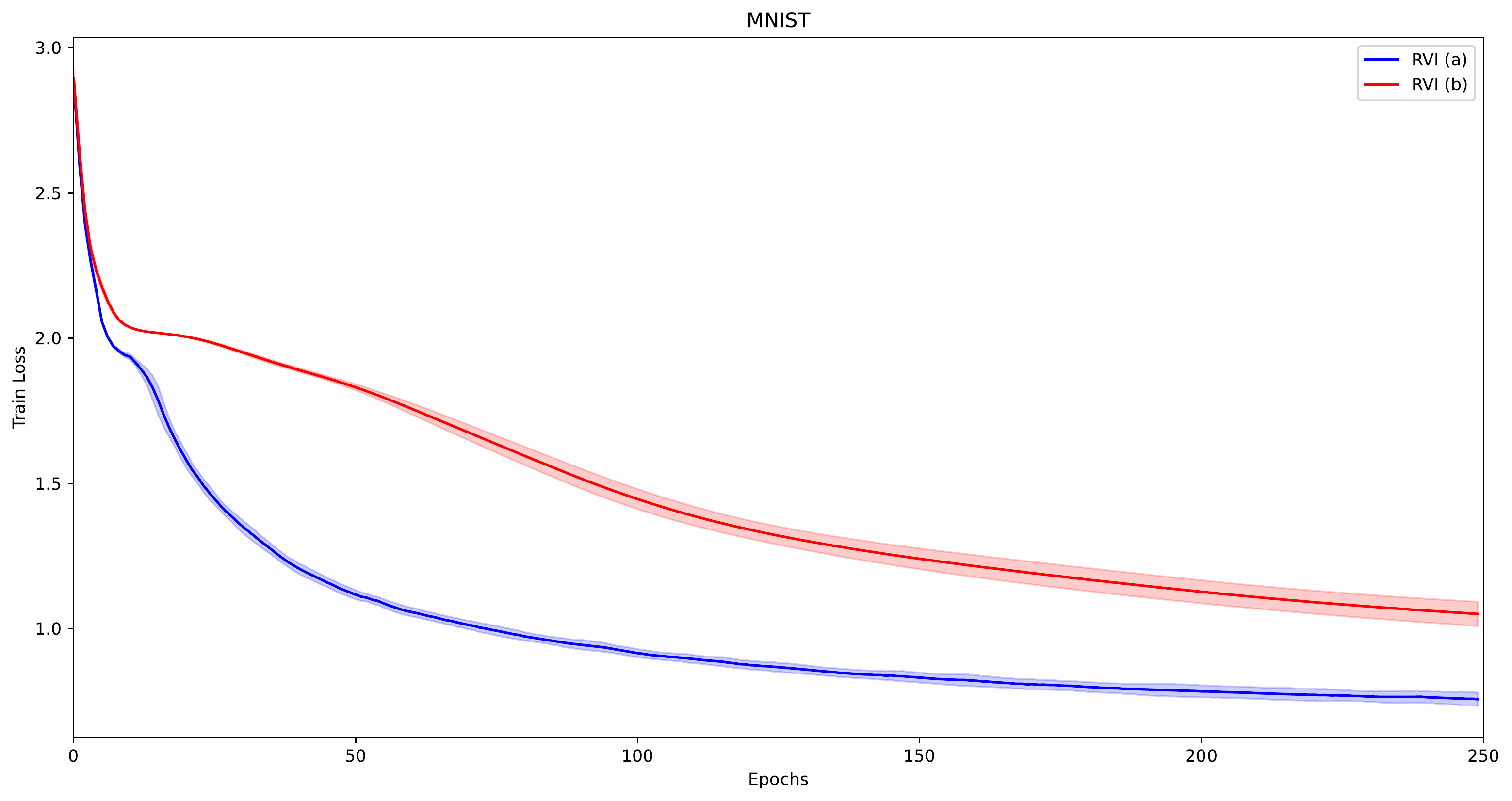}
\caption{\label{fig:boxes} Results for the experiments in Section \ref{app:formulation}. \ns \ (a) is with the clustering relay formulation and \ns \ (b) is the method used in the body of the paper. The latter shows faster convergence to a better loss. }}
\end{figure*} 

\section{Effect of $|B_i|$ on the Convergence of \ns}\label{app:frac}

\begin{figure*}[t]
\centering{
\includegraphics[width=\linewidth]{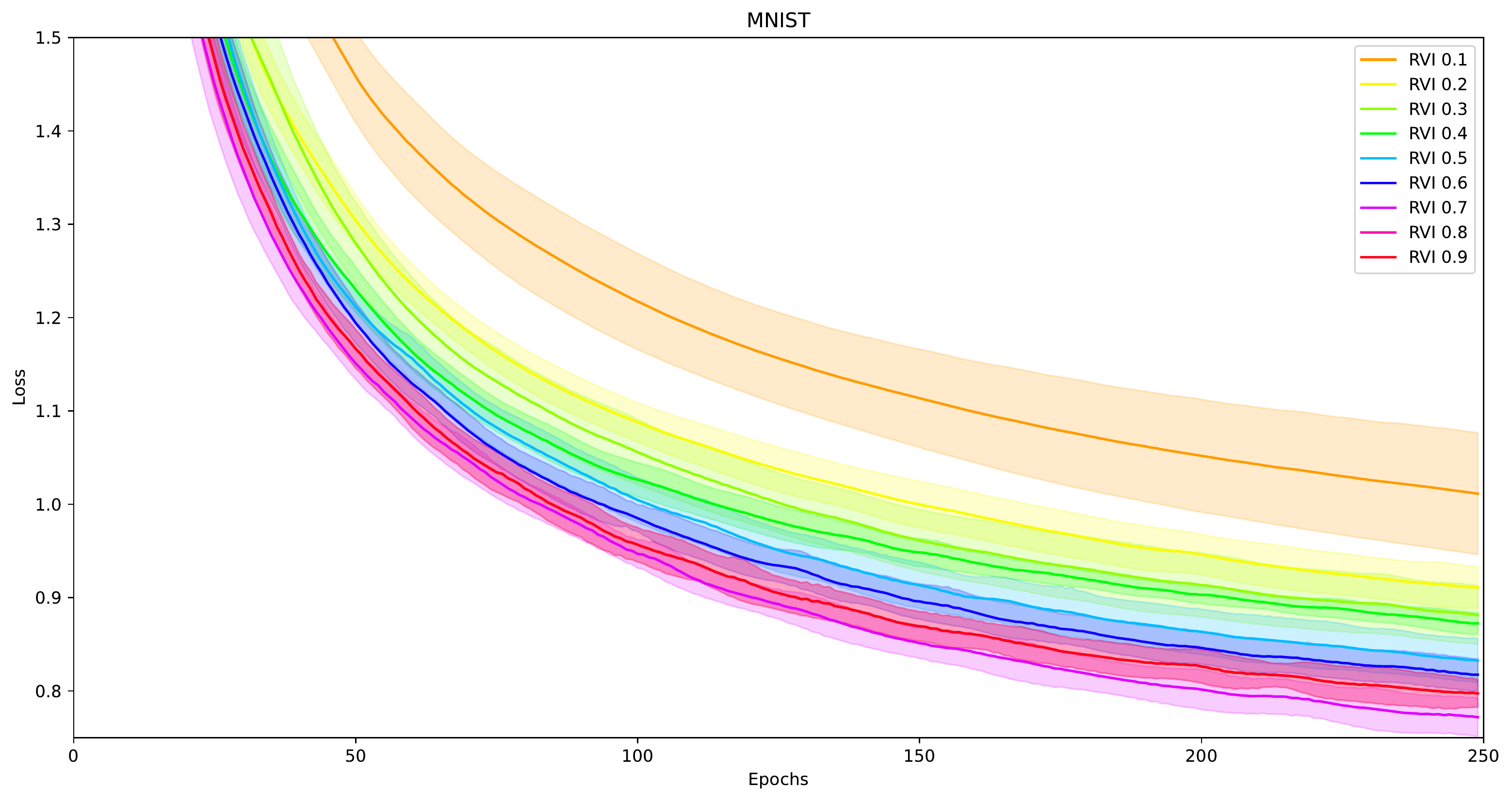}
\caption{\label{fig:cutoff} Convergence comparison for different ratios of $\frac{|B_i|}{K} \in \{0.1,0.2,0.3,0.4,0.5,0.6,0.7,0.8,0.9\}$ as discussed in Appendix \ref{app:frac}. Results are zoomed in to better show the distinction. }}
\end{figure*}

$B_i=\{b_i \in K\}$ essentially controlls how many shared vectors in the latent space each datapoint connects to, for the formulation in the main body of the paper. In our experiments, we change the proportion of $\frac{|B_i|}{K} \in \{0.1,0.2,0.3,0.4,0.5,0.6,0.7,0.8,0.9\}$. This essentially means each datapoint can connect (via coefficients $a_i$). We use the groupings of $\{25,50,100\}$ vectors. For example, for the case of 0.8, that would mean each datapoint can connect to $\{20,40,80\}$ from the original set of vectors. Results in Figure \ref{fig:cutoff} show the convergence for this experiment. MNIST with $0.5$ MCAR missing rate is used with learning rate of $0.001$ for all the learnable parameters of \ns. $\frac{|B_i|}{K}$ offers a tradeoff between performance in loss and performance in computational cost. However, we observe that $0.5$ is a suitable for outperforming VAE and VAD in most cases. 

\begin{figure*}[t]
\centering{
\includegraphics[width=.9\linewidth]{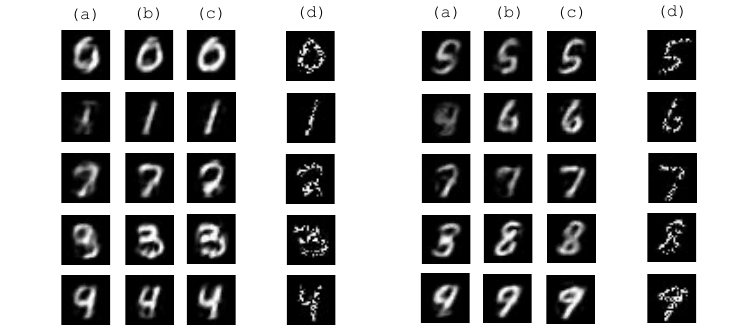}
\caption{\label{fig:config} Effect of relay configuration on the final reconstructions. Column $(\texttt{a})$ uses a single group of $[25]$ vectors, $(\texttt{b})$ uses double layers of $[25,50]$ and columns $(\texttt{c})$ uses groups of $[25,50,100]$ }}
\end{figure*}

In order to compare the effect of having different groupings at training time, we use three sets of groupings to train the \ns \ model: $\{25\}$, $\{25,50\}$ and c. Table \ref{fig:config} visually shows the training results. We observe that having different groups is beneficial for the final quality of reconstructions.  

\section{Parameterization of the $\phi_i$ and Calculation of ELBo}
In order to calculate the variational lower bound in Equation \ref{eq:elbo}, the approximate posteriors can be defined as a multivariate normal distribution:

\begin{equation}\label{eq:qnormal}
q_\phi(z|x_i) = \mathcal{N}(z; \mu^R_i+\mu^\epsilon_i,\Sigma_i)
\end{equation}

With the $\mu^R_i$ as the mean calculate via the relay formulation, and $\mu^\epsilon_i$ the individual deviations from the relay. $\Sigma_i$ is the standard deviation. Samples $z \sim q_\phi (\cdot)$ can then be defined as $z=\mu^R_i+\mu^\epsilon_i+\lambda\cdot \Sigma_i$, and $\lambda \sim \mathcal{N}(0,I)$. Subsequently, Equation \ref{eq:elbo} becomes differentiable to the samples drawn based on the Equation \ref{eq:qnormal}, essentially being differentiable w.r.t both $\phi^R_i$ and $\phi^\epsilon_i$. The final calculation depends on the choice for $p_\theta(x|z)$ - which in this paper is similar to \cite{zadeh2019variational}. The missing data will not partake in the calculation of the loss via marginalization. 

Similarly, after training is done, samples can be drawn using the same reparameterization trick, or alternatively sampling using the second term in the RHS of Equation \ref{eq:elbo}. 

\section{Further Training Details}
The training optimizer for all the experiments is Adam~\cite{kingma2014adam}. We set the batch size for the datasets at $256$. Due to heavy computational costs associated with the large number of experiments in this paper, we set the maximum data training size at random $10,000$ datapoints for the MNIST, FashionMNIST and Artificial datasets. In our experiments $10,000$ datapoints were enough to establish comparisons, as the results closely followed the full dataset experiments.

\section{Choice of Missing Patterns}
While in the main body we study the case of MCAR, we also study the case of $10$ and $20$ missing boxes of $4\times4$ pixels. Figure \ref{fig:boxes} shows the results of this experiment, with a similar trend as the MCAR. 

\begin{figure*}[t]
\centering{
\includegraphics[width=\linewidth]{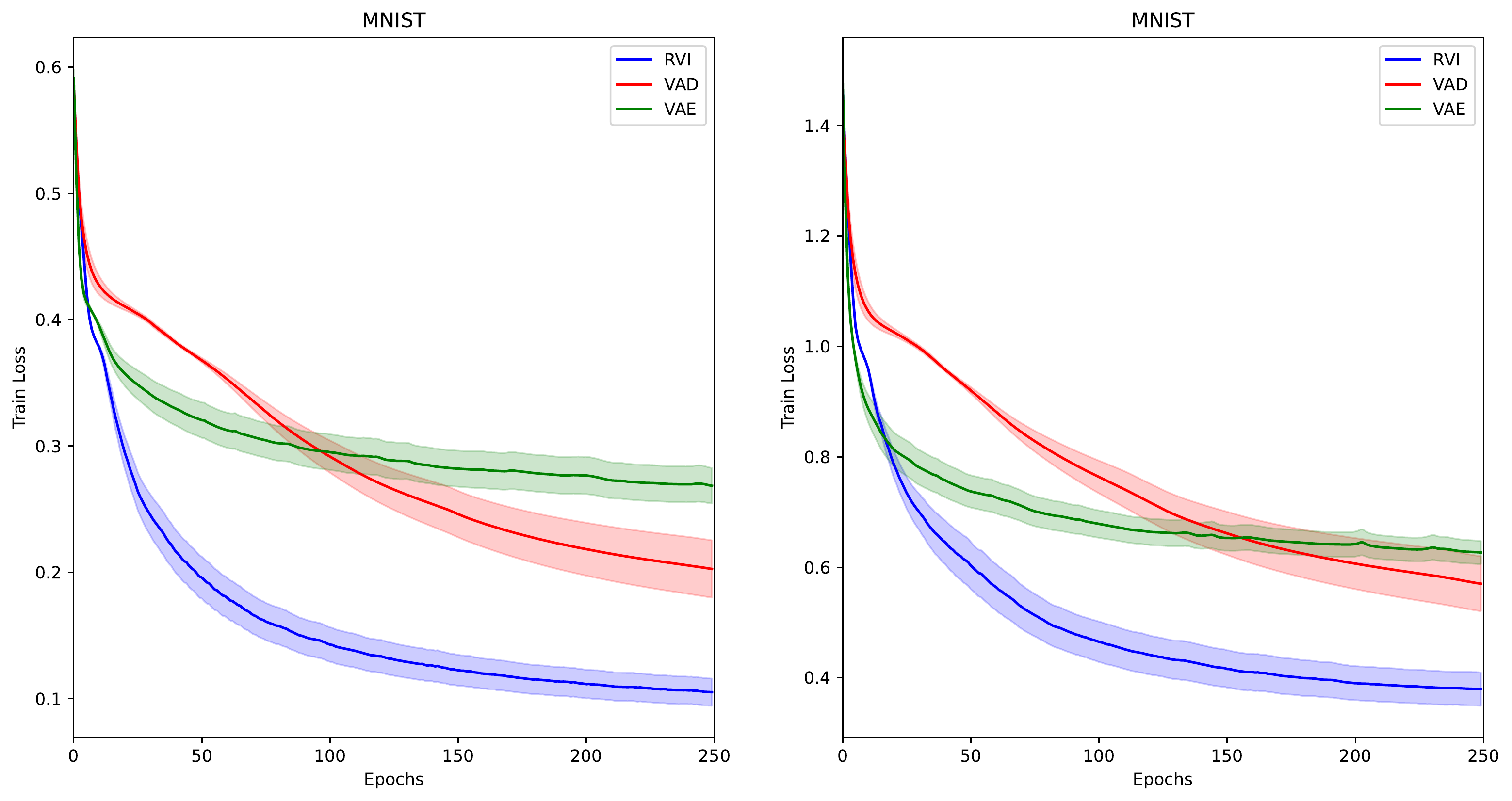}
\caption{\label{fig:boxes} Results for the boxes missing pattern. Right shows the $10$ missing, and left $20$ missing boxes of $4\times4$ pixels.}}
\end{figure*}

\section{Code}
Training code for RVI with all the formulations in Section \ref{app:formulation} are zipped and attached.

\end{appendices}
\end{document}